\documentclass[letterpaper, 10 pt, journal, twoside]{IEEEtran}

\IEEEoverridecommandlockouts                              

\usepackage{times} 
\usepackage{multicol}
\usepackage[bookmarks=true]{hyperref}
\usepackage{graphicx} 
\usepackage{subcaption}
\usepackage{epsfig} 
\usepackage{amsmath} 
\usepackage{amssymb}  
\usepackage{mathtools}
\usepackage{float}
\usepackage{multirow}
\usepackage{siunitx}
\usepackage{makecell}
\usepackage{hyperref}
\usepackage[dvipsnames]{xcolor}
\usepackage{pifont}
\usepackage{siunitx}

\newcommand\numberthis{\addtocounter{equation}{1}\tag{\theequation}}

\newcommand{\modif}[1]{{\textcolor{black}{#1}}}
\newcommand{\etal}{et al.}

\title{Neural Scene Representation for Locomotion on Structured Terrain}

\author{David Hoeller$^{1,2}$, Nikita Rudin$^{1,2}$, Christopher Choy$^{2}$, Animashree Anandkumar$^{2,3}$, Marco Hutter$^{1}$
\thanks{Manuscript received: February, 24, 2022; Revised: May, 15, 2022; Accepted: June, 06, 2022.}
\thanks{This paper was recommended for publication by Editor Jens Kober upon evaluation of the Associate Editor and Reviewers' comments.}
\thanks{This work was supported by NVIDIA, the Swiss National Science Foundation (SNSF) through project 188596, the National Centre of Competence in Research Robotics (NCCR Robotics), and the European Union's Horizon 2020 research and innovation program under grant agreement No.780883. Moreover, this work has been conducted as part of ANYmal Research, a community to advance legged robotics.}%
\thanks{$^{1}$ Affiliated with the Robotic Systems Lab, ETH Z\"u{}rich, Switzerland}%
\thanks{$^{2}$ Affiliated with NVIDIA}%
\thanks{$^{3}$ Affiliated with Caltech, USA}%
\thanks{Correspondence: {\tt\footnotesize dhoeller@ethz.ch}}
}

\begin{document}

\maketitle

\begin{abstract}
We propose a learning-based method to reconstruct the local terrain for locomotion with a mobile robot traversing urban environments. Using a stream of depth measurements from the onboard cameras and the robot's trajectory, the algorithm estimates the topography in the robot's vicinity.
The raw measurements from these cameras are noisy and only provide partial and occluded observations that in many cases do not show the terrain the robot stands on.
Therefore, we propose a 3D reconstruction model that faithfully reconstructs the scene, despite the noisy measurements and large amounts of missing data coming from the blind spots of the camera arrangement.
The model consists of a 4D fully convolutional network on point clouds that learns the geometric priors to complete the scene from the context and an auto-regressive feedback to leverage spatio-temporal consistency and use evidence from the past. 
The network can be solely trained with synthetic data, and due to extensive augmentation, it is robust in the real world, as shown in the validation on a quadrupedal robot, ANYmal, traversing challenging settings.
We run the pipeline on the robot's onboard low-power computer using an efficient sparse tensor implementation and show that the proposed method outperforms classical map representations.
\end{abstract}

\IEEEpeerreviewmaketitle


\section{INTRODUCTION}

\IEEEPARstart{H}{umans} and animals are endowed with a strong intuition about their physical environments. We rely on the spatial model we have formed from past experience to overcome common terrains with minimal mental effort. We can walk up a flight of stairs after a brief initial glance and can switch focus away from the stairs.
Similarly, getting an accurate terrain map from few scans is a crucial skill that robots need for locomotion and navigation, yet computing such a 3D map is a challenging problem due to noisy sensors, occlusions, drifts in localization, etc.~\cite{belter2012estimating,roennau2010six,Fankhauser2018ProbabilisticTerrainMapping,mastalli2017trajectory}.
Recent advances in deep learning have shown that it is also possible to train robust robotic systems from past data and synthesize the experience in a single neural network~\cite{schulman2017ppo}. In locomotion research, they have enabled unmanned ground vehicles such as legged robots to operate under harsh conditions \cite{miki2021locomotion}. 


However, walking on rough terrain is still challenging for quadrupedal robots. 
The robot has to estimate the terrain using exteroceptive sensors, and the locomotion controller needs to make sense of this high-dimensional information.
In many cases, exteroceptive sensors such as cameras and depth sensors suffer from large amounts of motion blur, changing lighting conditions, and occlusions. 
To make matters worse, commercially available quadrupeds such as ANYmal or Spot have a specific depth camera set-up that leaves blind spots in the critical areas for proper foothold placement, such as below the robot. 
Thus, mobile robots must first build a map using point cloud measurements and odometry information. 
The data is fused over time to estimate the terrain around the robot.
The surface is then passed on to the locomotion controller to choose the correct footholds to overcome the rough environment.
Due to the harsh conditions, heavy drifts in localization, and noisy depth observations, the reconstructed maps tend to be noisy and unfit for locomotion.
As a result, established methods such as Elevation Mapping~\cite{Fankhauser2018ProbabilisticTerrainMapping} or Voxblox~\cite{oleynikova2017voxblox} need to rely on many hyper-parameters and heuristics, requiring expert knowledge for tuning. In \cite{miki2021locomotion}, the authors address these issues and deploy a locomotion policy that internally updates the map estimate by combining the sparse height scans with proprioceptive sensing using a recurrent structure. However, the map representation is tied to the controller and cannot be used by another module. 



\begin{figure}[t]
    \centering
    \includegraphics[width=0.99\columnwidth]{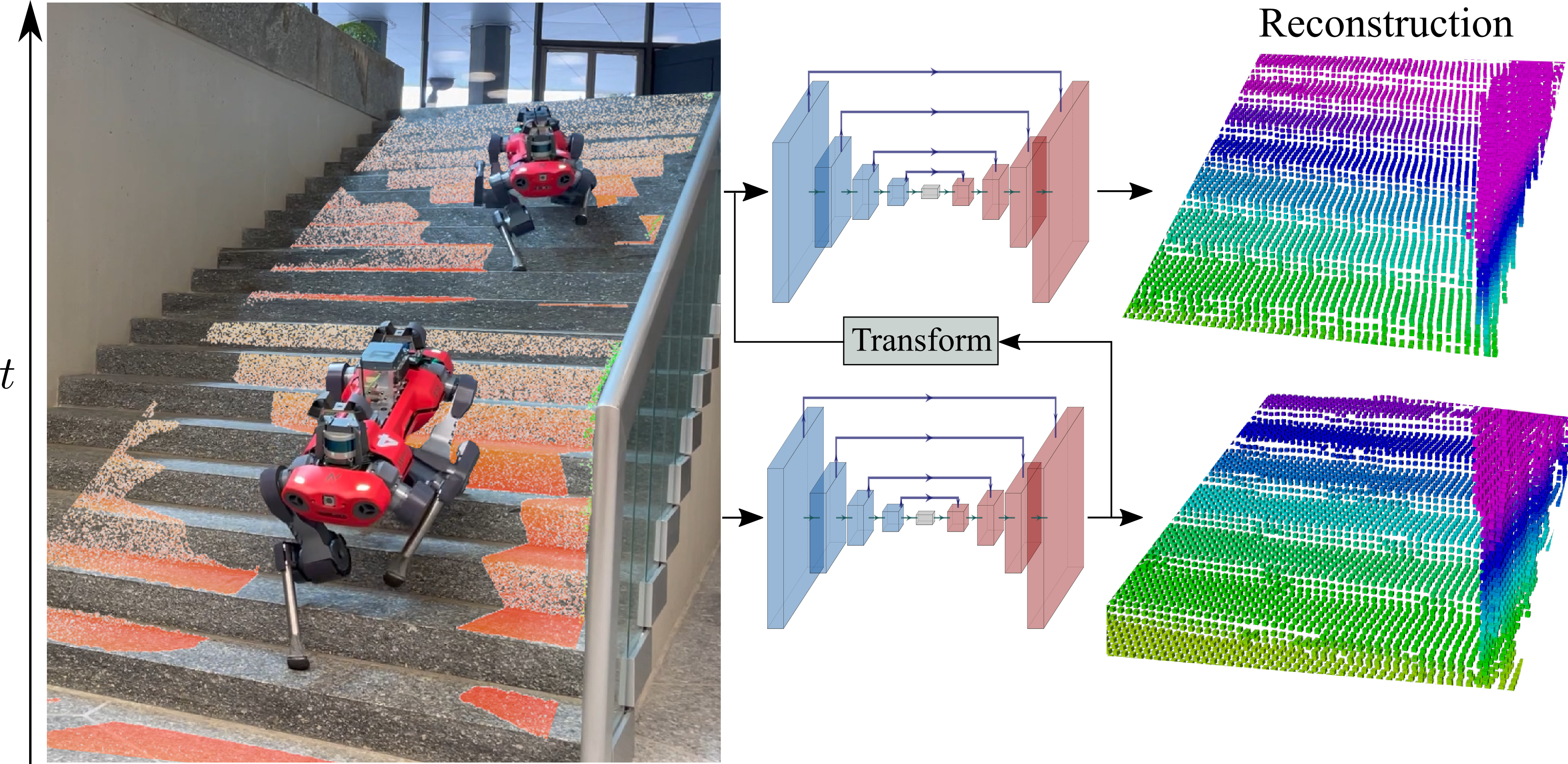}
    \caption{Overview of our approach. The noisy point cloud obtained from the depth sensors and the previous output are passed through the network, producing a point cloud estimate of the surrounding scene. 
    Due to the auto-regressive nature of the approach, the reconstruction is refined over time, and the network remembers objects that have entered the robot's blind spots.}
    \label{fig:pipeline}
    \vspace{-4mm}
\end{figure}  

To tackle these challenges, we present a learning-based algorithm that estimates the local terrain around the robot from a stream of noisy depth measurements, see Fig.~\ref{fig:pipeline}.
The pipeline takes the robot's pose, the partial observations from the cameras, and the latest map estimate as input and reconstructs the scene, even in the occluded regions.
The auto-regressive feedback allows the network to maintain the detailed geometry using the information from the previous inputs.
We train the network to build an intuition about the spatial configuration of the scene from the current context, similar to what humans do.
When walking on stairs, for example, despite only observing the steps partially, the module understands the situation and extends the stairs below the robot, even when a large portion of the surface is occluded.
Our network also uses the temporal sequence to memorize completely occluded objects that could be seen in past measurements, such as when walking over a box.

We use NVIDIA's IsaacGym \cite{IsaacGym} simulation environment to create a large number of randomized scenes with walls, roadblocks, flights of stairs, and boxes. Our robotic agents move around the scene to collect a data-set consisting of more than 200,000 point cloud observations.
During training, we simulate the depth camera noise and apply extensive data augmentations making the 3D reconstruction network predict the ground-truth environment despite excessive noise and incomplete observations.
Our module is trained on synthetic data only, and due to the augmentations, it transfers sim-to-real and generalizes to real-world sensory data, as we show in our experiments in various environments.









While we do not claim to solve the complete 3D rough terrain problem and focus on the simpler urban setting with structure, this is a promising step towards that direction. 
To summarize, this paper presents the following contributions:
\begin{itemize} 
\item A novel approach that reconstructs urban-like environments under the harsh requirements of the real world, i.e. noisy point clouds, missing data from stereo matching failures, noisy state estimation from jerky motions and impacts with the world resulting in imprecise trajectory estimates.
\item A method to initialize the map with a meaningful guess without heuristics using the context inferred from the visible data.
\item An evaluation in simulation and on the real robot, showing that the approach can handle state estimator drifts and large amounts of missing data from readily-available depth cameras and outperforms currently used baselines that rely on heuristics.
\end{itemize}

\section{RELATED WORK}

\begin{figure*}[t!]
\centering
\vspace{4mm}
\includegraphics[width=0.9\textwidth]{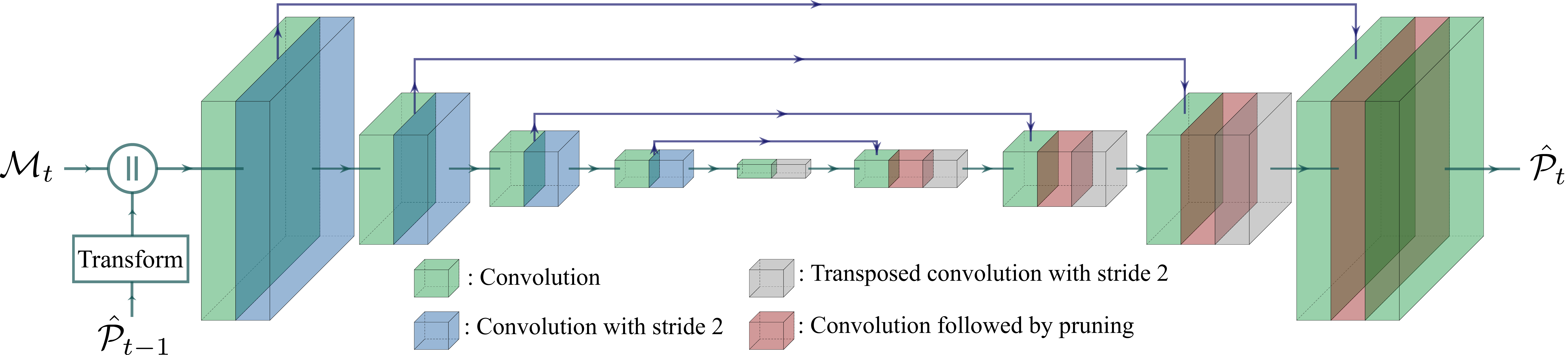}
\caption{Architecture of the terrain representation module. The previous point cloud estimate $\hat{\mathcal{P}}_{t-1}$ at time $t-1$ is first transformed into the current frame and concatenated temporally with the current point cloud measurement $\mathcal{M}_{t}$ at time $t$. The result is fed through a fully convolutional encoder-decoder network with skip connections to produce an estimate of the current point cloud $\hat{\mathcal{P}}_{t}$ at time $t$.}
\label{fig:architecture}
\vspace{-4mm}
\end{figure*}

\paragraph{Locomotion}
Legged locomotion is a well-studied field of robotics. 
While past works mainly focus on model-based control \cite{kim2019loco}, learning-based methods have increasingly come to the spotlight due to recent advances in deep reinforcement learning \cite{schulman2017ppo}. 
Blind quadrupedal locomotion is on flat terrain using a neural network is demonstrated in \cite{tan2018simtoreal} and \cite{hwangbo2019RL}. 
The authors in \cite{lee2020locomotion} and \cite{siekmann2021Cassie} build upon these works and show that it is to some extent possible to walk on rough terrain without exteroceptive sensors with a quadruped and a biped, respectively. 
Newer approaches show successful locomotion on more challenging terrain using perceptive inputs \cite{miki2021locomotion, rudin2021learning, gangapurwala2020RLOCTL}. A carefully designed architecture is used in \cite{miki2021locomotion} to switch between a blind controller and a perceptive one depending on the quality of the terrain reconstruction and heavily noisify the terrain measurements during training. 
\modif{In \cite{yang2022learning}, the policy uses a transformer to process proprioceptive and exteroceptive data to perform locomotion and obstacle avoidance. However, similar to \cite{miki2021locomotion}, the vision model is tied to the policy and has to be retrained for every new policy.
In the experiments, we show that our method works with cheap and readily available cameras and that is can be used as a standalone module by different model-based and learning-based policies without retraining.}

\paragraph{Environment Mapping}
Mapping the 3D environment using exteroceptive sensing has been studied in the context of robotics. The three most commonly used representations are 2.5D elevation maps \cite{belter2012estimating,roennau2010six,mastalli2017trajectory,Fankhauser2018ProbabilisticTerrainMapping}, point clouds \cite{klein2007parallel, forster2014svo, engel2014lsd}, and voxel grids \cite{hornung13octomap, oleynikova2017voxblox}.

The elevation map representation is widely used in robotics for locomotion \cite{kim2020loco, miki2021locomotion, rudin2021learning} and planing \cite{chavez2018traversability} due to its simplicity. The terrain is encoded as a top-down view image around the robot.
In \cite{Fankhauser2018ProbabilisticTerrainMapping}, the authors use a probabilistic formulation and fuse the range measurements and the robot's pose using a Kalman filter to produce an estimate of the height profile of the surrounding terrain. The main drawback is that it is susceptible to drift in odometry. Also, height maps cannot represent the scene in full 3D, and a table, for example, is represented as one single block. The authors in \cite{9676411} extend the approach and perform hole filling on the height map with a neural network. The method does not perform filtering in the visible regions and since it only considers the current time step, it also suffers from drifting issues. 


Point cloud-based methods are mainly used in the context of Simultaneous Localization and Mapping (SLAM). Such systems use RGB or range information to produce an estimate of the robot's trajectory and incrementally build a map of the world. PTAM~\cite{klein2007parallel} is a lightweight parallel mapping and tracking algorithm that uses a sparse set of image correspondences.
Semi direct~\cite{forster2014svo} and direct~\cite{engel2014lsd} odometry frameworks use photometric errors to map the environment densely and estimate the relative poses instead of feature-based matching. 


Lastly, voxelized-based mapping has also been used for 3D reconstruction. OctoMap~\cite{hornung13octomap} is an octree-based hierarchical probabilistic 3D voxel environment representation. This representation is efficient and fast, but only uses voxel-wise binary occupancy. 
Voxblox~\cite{oleynikova2017voxblox} is a similar volumetric mapping library that uses voxel hashing~\cite{niessner2013real} to grow the environment dynamically and saves distance to the closest surface in the form of a Euclidean Distance Transforms (EDT).

Most of these 3D mapping algorithms rely on classical methods to reconstruct the scene and thus lack the capabilities to semantically complete missing information. 
We use a voxel-based approach, where each voxel stores the relative coordinates of the corresponding point to achieve sub-voxel accuracy. Unlike previous works, we use a learning-based algorithm to complete unseen parts of the scene using a neural network and handle drift in odometry.

\paragraph{3D Scene Completion}

In many cases, commercial 3D scanners fail to reconstruct the geometry of the scene accurately due to various factors such as registration failure and occlusion. Such incomplete 3D scans can cause errors in the subsequent parts of the pipeline and thus, many recent works have proposed learning-based 3D scene completion from partial/incomplete 3D scans. Song~\etal~\cite{song2017semantic} proposed voxelized dense 3D scene completion with semantic labels. Dai~\etal~\cite{dai2018scancomplete} use similar dense 3D voxelized 3D completion but add an auto-regressive hierarchical structure to create a high-resolution completion. \modif{In VolumeFusion \cite{Choe2021VolumeFusion}, the authors take a sequence of RGB images to estimate the depth maps and fuses them with the images' features to estimate the truncated signed distance function of the scene, but has no proper recurrence to keep information from the past.} Sun~\etal~\cite{sun2021neuralrecon} propose a 3D reconstruction network that completes a scene using a hierarchical spatially sparse neural network. The network up-samples voxels and prunes unnecessary ones to reduce the computational cost of high-resolution reconstruction. This is similar to Gwak~\etal~\cite{gwak2020gsdn} who employ generative convolutions and pruning. We adopt a similar approach of up-sampling and pruning voxels to generate high-resolution 3D reconstructions while reducing the computational complexity.


\section{METHODOLOGY}

This section describes how we reconstruct the terrain from noisy and occluded observations. The pipeline is depicted in Fig.~\ref{fig:architecture}. Using the pose difference between the previous time step and the current one, the previous point cloud estimate $\hat{\mathcal{P}}_{t-1}$ (output of the network) is first transformed to the current measurement frame. The result is concatenated with the current measurement $\mathcal{M}_{t}$, and fed into a fully convolutional network, producing the point cloud estimate $\hat{\mathcal{P}}_{t}$.


\subsection{Input Pre-processing}
First, the point cloud has to be converted to a data structure that can be forwarded through the network. We choose a voxel grid representation, where the voxels store the relative coordinates of the points to achieve sub-voxel accuracy. Since a dense grid representation has cubic complexity and could take up a large amount of memory, we use a sparse formulation.

%

To achieve this, we discretize the current point cloud measurements into a 64$\times$64$\times$64 grid that represents a \SI{3.2}{\meter}$\times$\SI{3.2}{\meter}$\times$\SI{3.2}{\meter} map around the robot.
For each voxel, we define a feature as the offset between the centroid of the points that fall within that voxel and the voxel's bottom left rear corner. Mathematically, let ${\mathbf{p}_i = [x_i, y_i, z_i] \in \mathbb{R}^3}$ be the centroid point at the $i$-th occupied voxel \modif{in the grid's reference frame, i.e. one unit is equal to one cell. The sparse input tensor in coordinate list format (COO) maps the discretized cell coordinate $c_i$ of that centroid to a feature value $f_i$, which are defined as}
\begin{align*}
c_i &= [\lfloor x_i \rfloor, \lfloor  y_i \rfloor, \lfloor z_i \rfloor, k_i], \\
f_i &= \mathbf{p}_i \; \mathrm{mod} \; 1 \numberthis \label{eq:voxel}
\end{align*}

\modif{where $\mathrm{mod}$ is the modulo operator, and $k$ is the discrete time index of that point. $k$ is set to 0 for the points of the current measurement and to 1 for the points from the previous output. This introduces a temporal component to the problem and lets the network perform 4D convolutions across space and time.}
\modif{Due to the modulo operation, the feature values represent the normalized coordinates of the centroid in each voxel and are in the range $[0, 1]$. The continuous centroid location can be retried by adding the cell index with the corresponding feature value.}




\subsection{Architecture}


The network is a \modif{U-Net-like \cite{ronneberger2015Unet}} fully convolutional 4D encoder-decoder network with skip connections. This architecture allows the decoder to maintain the fine details from the encoder.


The encoder is a sequence of convolutions that down-sample the input by a factor of 16 for each spatial axis. It uses 4 strided convolutions, each down-sampling the coordinates by a factor of 2 spatially but preserving the temporal dimension. As a result, the network keeps the two temporal channels separate and performs convolutions across space and time at the different resolutions.


The decoder uses the latent tensor as input. Additionally, the feature maps of each encoder block are forwarded to the corresponding decoder block using skip connections. The sequence of convolutions up-sample the tensors back to the same stride as the input. However, since the latent tensor is likely to be a fully occupied block of dimension 4x4x4x2, standard up-sampling would produce a fully occupied voxel grid with 64x64x64x1 cells. This would become prohibitively slow and would thwart the training process due to the sparsity of the problem.
Therefore, similar to \cite{gwak2020gsdn}, we introduce pruning operations in the decoder. This step discards elements from the feature maps before passing the data to subsequent stages. Specifically, at each layer, we take the features from the current sparse tensor $T_f$ to generate a likelihood for each voxel using a separate convolutional layer and a sigmoid activation, resulting in a sparse tensor $T_p$. We prune all the elements of $T_f$ whose likelihood in $T_p$ is smaller than a constant $\alpha \in [0,1]$. The tensor $T_p$ is not forwarded to the next layers.

The final convolution of the decoder maps the features to a dimension of 3 to produce an estimate of the sub-voxel position of the points, as in Eq.~\ref{eq:voxel}. 

\subsection{Training Procedure}
The target for reconstruction is the ground truth point cloud transformed according to Eq.~\ref{eq:voxel}. The loss function is composed of two parts. The first one computes the binary cross-entropy between the features of $T_p$ at each layer and a value of 1 or 0, depending on whether an occupied voxel is present at these coordinates in the target at the resolution of that layer. The other part takes the mean Euclidean distance between the output features and the features of the target to estimate the sub-voxel position of the points. Due to that formulation, at a given coordinate of $T_p$, the feature value represents the network's confidence that the voxel is occupied. 


The pruning threshold $\alpha$ is a key hyper-parameter that can be modified to change the behavior of the network. When $\alpha$ is high, the network is conservative and reconstructs only the points it is confident about. However, this reduces the output density, and holes can appear in the reconstruction. When $\alpha$ is reduced, the output becomes denser, but the network generates points not necessarily in the target. We found experimentally that setting $\alpha$ to 0.5 results in a good trade-off between the estimate's recall and precision. 

\modif{To train the network, we roll it out on a sequence of 12 time steps and compute the loss at each step. We found experimentally that propagating the gradients over time did not provide any benefit on reconstruction performance. We use the Adam optimizer with an initial learning rate of 0.01 that is exponentially decayed until 0.0001.}

\begin{figure}[t]
    \vspace{2mm}
    \centering
    \includegraphics[width=0.48\columnwidth]{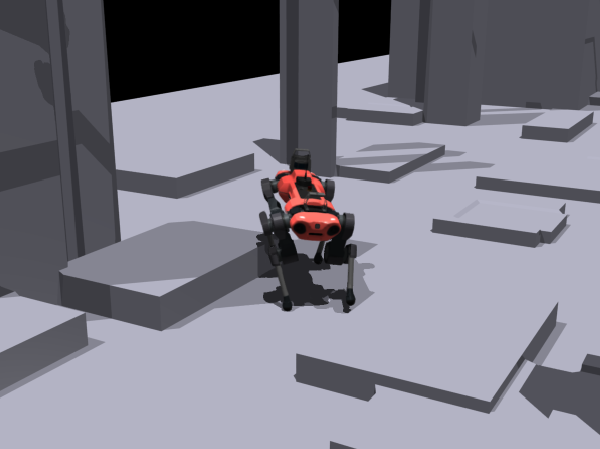}
    \includegraphics[width=0.48\columnwidth]{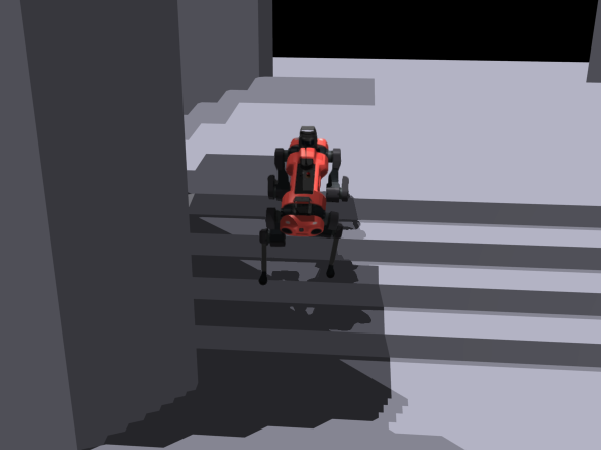}
    \caption{Randomized environments generated in simulation. \modif{The parameters of the scene are sampled uniformly: the stairs' width and height between $[0.2, 0.5]$\SI{}{\meter} and $[0.08, 0.25]$\SI{}{\meter}, respectively; the boxes' width and length between $[0.2, 2.0]$\SI{}{\meter} and their height between $[0.08, 0.25]$\SI{}{\meter}. The walls are sampled to produce corridors of width in the the range $[2, 6]$\SI{}{\meter}.}}
    \label{fig:environment}
    \vspace{-4mm}
\end{figure}

\subsection{Data Generation}
The architecture described in the previous subsection can estimate the ground truth point cloud around the robot from the stream of noisy point cloud measurements. To further encourage the module to reconstruct scenes with structure, we collect a large data set in urban-like settings. As a result, the network overfits to such environments and tries to implicitly identify key parameters of the scene, such as the length and height of stairs from the noisy data.

We generate structured environments consisting of stairs, boxes on the ground, walls, poles, and narrow corridors in simulation (Fig.~\ref{fig:environment}) using NVIDIA's IsaacGym \cite{IsaacGym}. The parameters of the scene's elements, such as the boxes' dimensions or the walls' locations, are randomized to generate a data set that reflects the real world's diversity.

The robot is controlled towards a reachable position with a randomized speed and base orientation using a rough terrain locomotion policy. 
\modif{The measurements are provided by four simulated Intel RealSense depth cameras placed at the front, back, left, right of the robot and tilted downwards by \SI{30}{\degree}, which is the standard configuration on the ANYmal C robot}. The ground truth point clouds are obtained by sampling a dense point cloud from the mesh of the terrain around the robot at every time step. \modif{On average, 43\% of the ground truth points are visible in the measurements in the data-set comprising of 200000 time steps.}

\subsection{Data augmentation}
During training, the measurements are noisified (Fig.~\ref{fig:results_rec_sim}) to make the pipeline robust to the noise of the real system and facilitate sim-to-real transfer. We using the following augmentations:
\begin{itemize}
    \item \textit{Position}: The position of each point is disturbed uniformly in the range $[\SI{-0.05}{\meter}, \SI{0.05}{\meter}]$
     \item \textit{Tilt}: The point cloud is tilted in a random direction by an angle sampled uniformly in the range $[\SI{-1}{\degree}, \SI{1}{\degree}]$   
    \item \textit{Height}: The height of random patches of the point cloud is disturbed uniformly in the range $[\SI{-0.05}{\meter}, \SI{0.05}{\meter}]$
    \item \textit{Pruning}: Random patches of the point cloud are removed
    \item \textit{Outliers}: Random clusters of points are added to the measurement
    \item \textit{Robot Pose}: The position of the robot is uniformly disturbed in the range $[\SI{-0.05}{\meter}, \SI{0.05}{\meter}]$
\end{itemize}
We also randomly mirror the data along the x and y axes for a whole trajectory to produce more diverse trajectories.

The network is therefore trained for denoising and completion and has to utilize the evidence from previous measurements to estimate the partially observable state of the world. 

The backbone of our pipeline relies on the Minkowski Engine~\cite{choy20194d}, which provides CPU and GPU accelerations for neural networks for spatially sparse tensors.

\section{EXPERIMENTS}

We validate our approach with various experiments in simulation and on the real robot and compare against the baselines Elevation Mapping \cite{Fankhauser2018ProbabilisticTerrainMapping} and Voxblox \cite{oleynikova2017voxblox}. While both baselines have more than ten hyper-parameters and heuristics, our approach only has the pruning threshold $\alpha$ that needs tuning. We report the mean precision, recall, F1 score by discretizing the world in a robot-centric voxel grid of dimension 64$\times$64$\times$64 with cell sizes \SI{0.05}{\meter}$\times$\SI{0.05}{\meter}$\times$\SI{0.05}{\meter}, as well as the mean the absolute height difference between the reconstruction and the ground truth. The latter is essential for the locomotion task since the policy directly uses the height information.

\subsection{Evaluation in Simulation}

\begin{figure}[t]
    \centering
    \vspace{2mm}
    \begin{subfigure}[b]{\columnwidth}
    \vspace*{1mm}
        \centering
        \includegraphics[width=0.48\columnwidth]{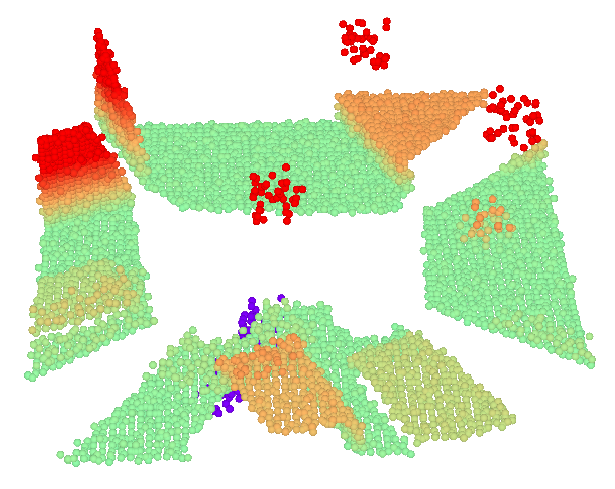}
        \includegraphics[width=0.48\columnwidth]{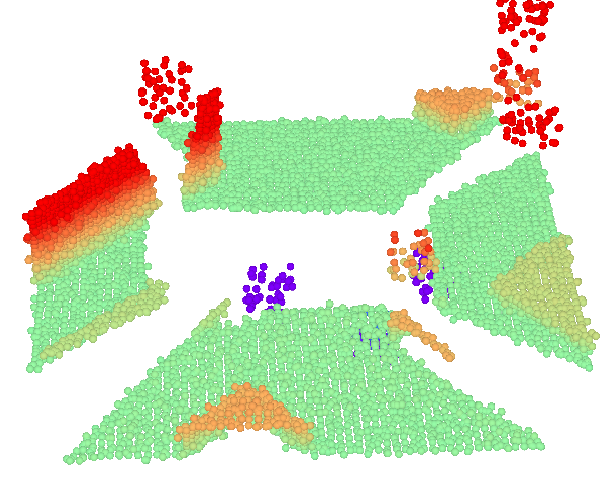}
        \caption{Measurement.}
    \end{subfigure}
    \begin{subfigure}[b]{\columnwidth}
    \vspace*{1mm}
        \centering
        \includegraphics[width=0.48\columnwidth]{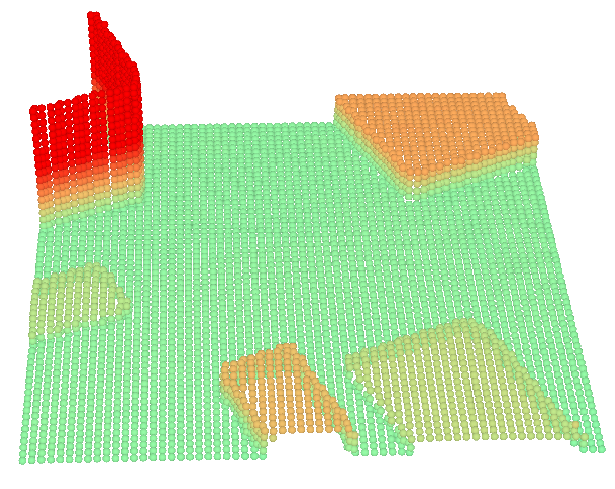}
        \includegraphics[width=0.48\columnwidth]{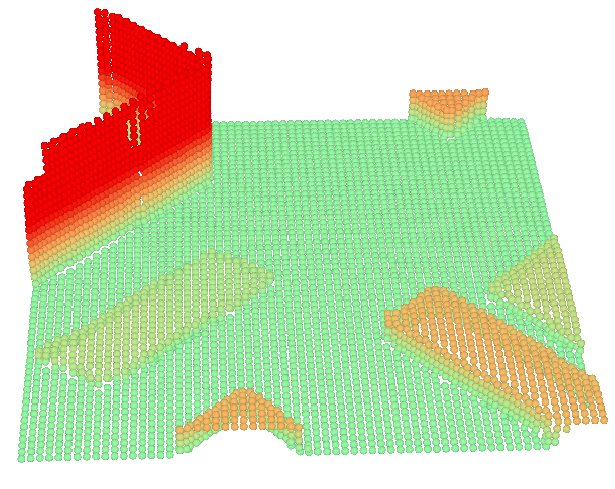}
        \caption{Reconstruction.}
     \end{subfigure}
    \begin{subfigure}[b]{\columnwidth}
     \vspace*{1mm}
        \centering
        \includegraphics[width=0.48\columnwidth]{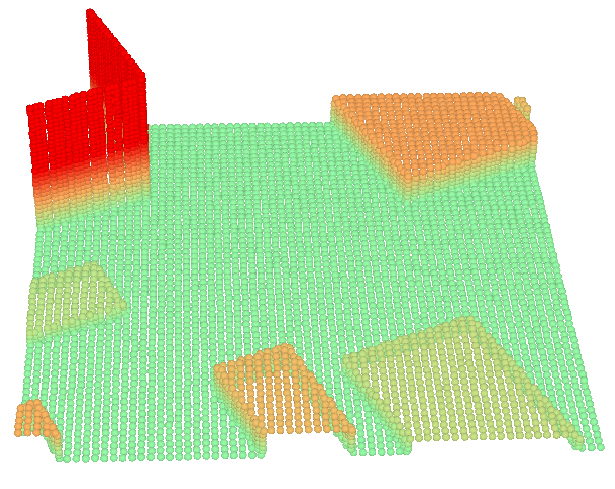}
        \includegraphics[width=0.48\columnwidth]{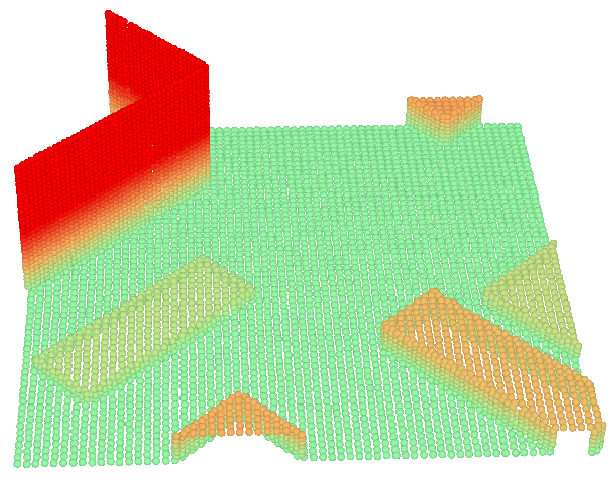}
        \caption{Ground truth.}
    \end{subfigure}
    \caption{Reconstruction on a validation trajectory in simulation. The left column shows the estimate at the beginning of the trajectory, and the right the estimate \SI{1.5}{\second} later. While two boxes in the diagonals cannot be seen anymore in the measurements on the right, they are still reconstructed correctly using the evidence from the past.}
    \label{fig:results_rec_sim}
    \vspace{-4mm}
\end{figure}

\begin{figure}[t]
    \centering
    \vspace{2mm}
    \begin{subfigure}[b]{0.48\columnwidth}
    \vspace*{1mm}
        \centering
        \includegraphics[width=\columnwidth]{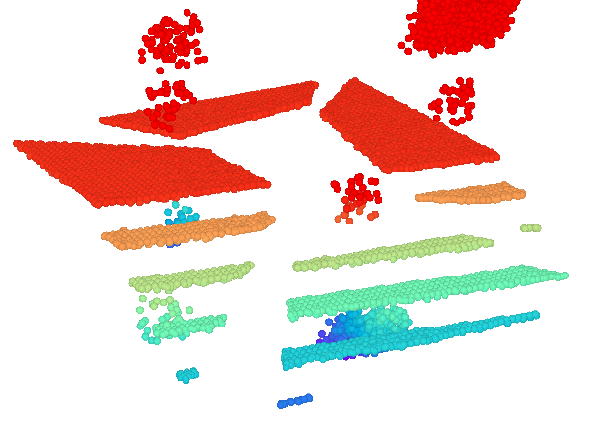}
        \caption{Measurement.}
    \end{subfigure}
    \begin{subfigure}[b]{0.48\columnwidth}
    \vspace*{1mm}
        \centering
        \includegraphics[width=\columnwidth]{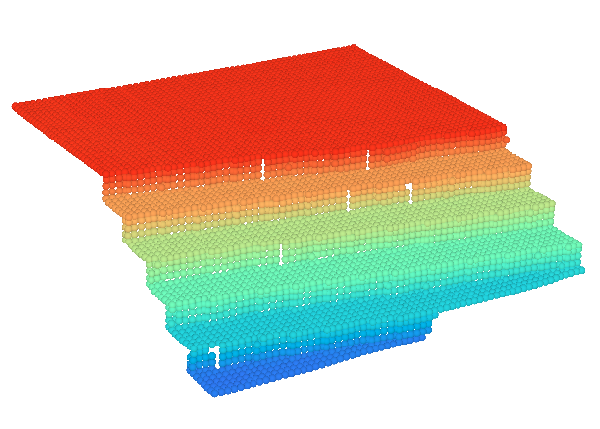}
        \caption{Reconstruction.}
     \end{subfigure}
    \caption{Zero-shot estimation on stairs, i.e., using only the measurement at the current time-step. The approach can correctly reconstruct the vertical surfaces of the walls using spatial data only.}
    \label{fig:results_zs_sim}
    
\end{figure}

The output of our module on a validation trajectory is depicted in Fig.~\ref{fig:results_rec_sim}. The left column shows the reconstruction at the beginning of a trajectory, while the right one the reconstruction \SI{1.5}{\second} later. Since the boxes could be seen in previous measurements, they are properly reconstructed on the right, despite being in the blind spots. It can also be seen that the approach correctly recreates the wall despite only seeing the bottom of it. 
This is because all the walls have the same height in the training data. This choice was made to reduce the reconstruction to only the relevant features for locomotion. Indeed, the robot cannot overcome larger obstacles, and representing the correct height of such elements does not bring any benefit.
Fig.~\ref{fig:results_zs_sim} shows the zero-shot estimate of the network on stairs, meaning that the output is computed using only the current measurement. We can conclude that in such scenes, the network can understand the context from spatial data only and generate the correct structure, e.g., the vertical surfaces of the stairs. Of course, adding the temporal information is necessary to reconstruct elements such as those in the left column. 

\begin{figure}[t]
    \centering
    \includegraphics[width=0.99\columnwidth]{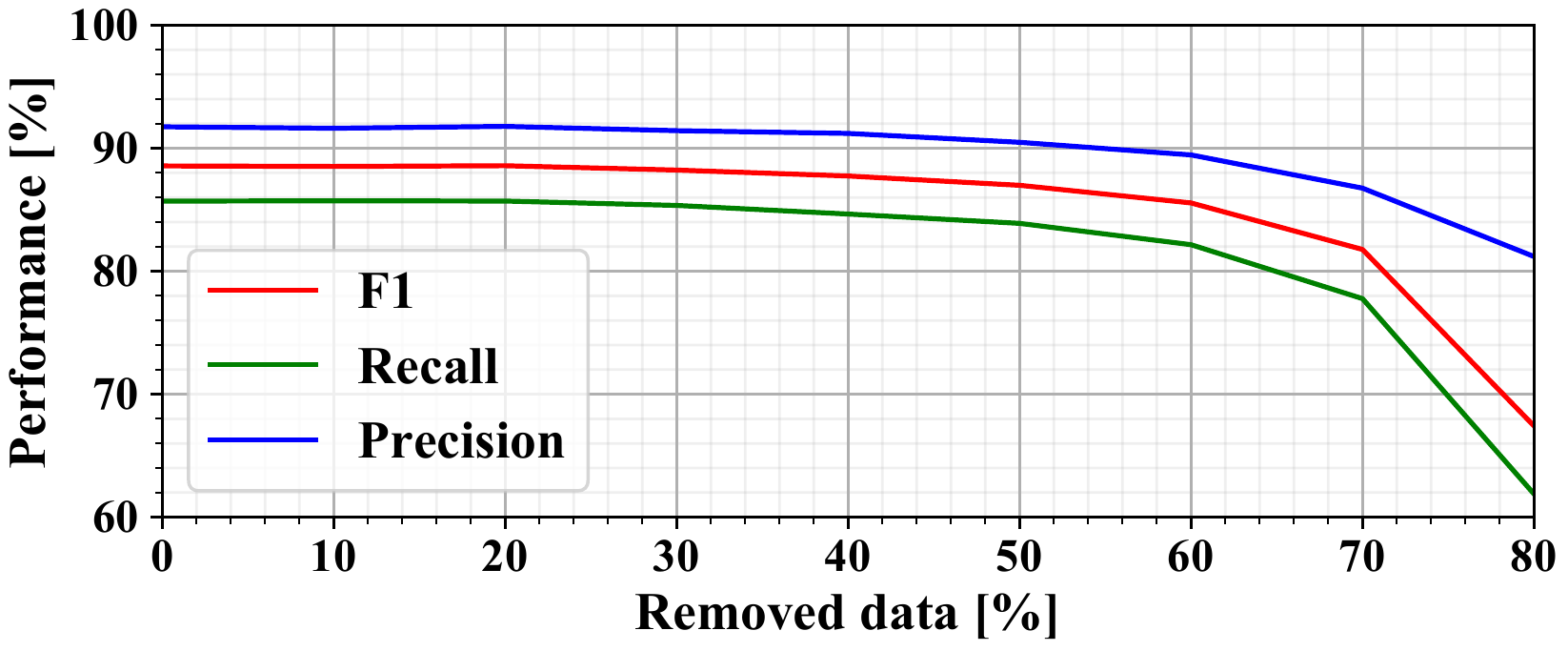}
    \caption{Mean performance on validation trajectories as a function of the amount of data removed from the measurements. The performance stays constant up to 50\% of data removal, after which the recall diminishes rapidly.}
    \label{fig:result_prune}
    \vspace{-4mm}
\end{figure}

We further analyze the output of the module as a function of the measurement density (Fig.~\ref{fig:result_prune}). We assess the performance for different amounts of data removed from the measurements for the same validation trajectories. Note that the reported removal rate does not consider the already missing points from the blind spots. The network can cope with up to 50\% of data omission, with an F1 score of 88\%. The output density then decreases sharply and holes appear, which is reflected by the recall decreasing to a value of 62\%. On the other hand, the precision does not decline as much and reaches a value of 82\%. When more than 80\% of data is omitted, the decoder cannot handle the sparsity and the pruning produces empty tensors. The results show that the network can handle sparse measurements and reconstruct the terrain accurately. It aggregates data on a very local level to produce a bigger picture of the scene. We hypothesize that this is due to the spatial formulation of the problem, where the 3D coordinates of the points in Eq.~\ref{eq:voxel} induce a strong prior about the configuration of the scene. This is in contrast to 2D-based methods on depth images, where the depth information is encoded in the input features. The complete 3D spatial information is not conserved in the sense that a convolution between two neighboring pixels always occurs, regardless of whether the corresponding points in 3D are very far apart or not.

\subsection{Evaluation on the Robot}
The network is deployed on an NVIDIA Jetson Xavier on the robot. A ROS node processes the incoming point cloud measurements \modif{coming from the Intel RealSense cameras at the front, left, right, and back of the robot} on a separate thread and maps the data to the GPU for inference. \modif{The pose is taken from the state estimator running on the robot.} The node then publishes the estimated point cloud, which the locomotion controller uses to query an array of heights around the robot for the policy. Note that during all the experiments, the locomotion controller uses the output of our module for terrain sensing, which shows that it can be deployed in real-world scenarios indoors and outdoors.

Inference on the on-board computer takes on average \SI{70}{\milli \second}, and the whole node, including the point cloud conversions, inference, and publishing runs at \SI{6}{\hertz}. \modif{Due to the computational limits of the hardware, we discard the measurements in between map updates. The reconstructed map around the robot has a dimension of $3.2\times3.2\times3.2 $ \SI{}{\meter}, while the terrain input for the locomotion policy is $1.6\times1.0$ \SI{}{\meter}. Since the controller runs faster at a rate of \SI{50}{\hertz}, the perceptive inputs for the policy are computed at the pose relative to the latest map. As a result, the robot could walk with a speed of up to \SI{4.8}{\meter / \second} before reaching the limit of the map. This is more than enough since we run our policy at a maximum speed of \SI{1}{\meter / \second}.}

\begin{figure}[t]
    \centering
    \begin{subfigure}[b]{\columnwidth}
        \centering
        \includegraphics[width=0.4\columnwidth]{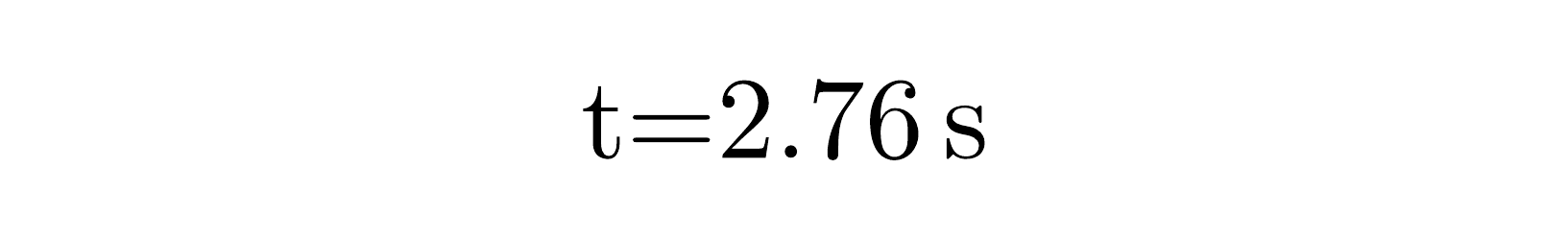}
        \hspace{0.1\columnwidth}
        \includegraphics[width=0.4\columnwidth]{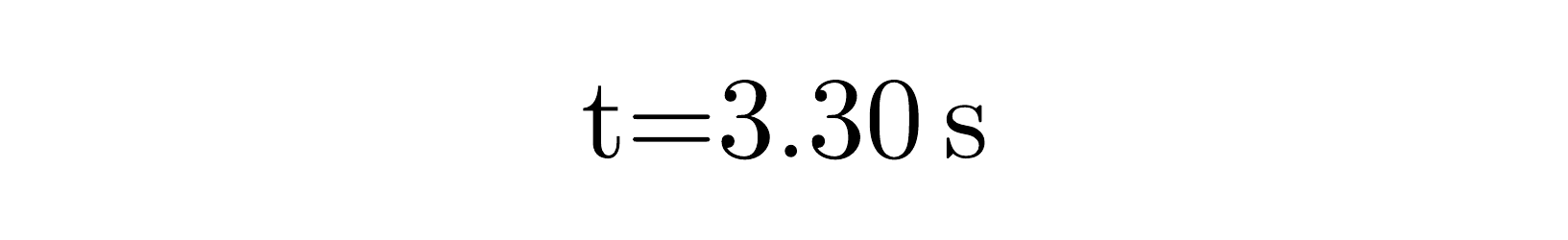}
    \end{subfigure}
    \begin{subfigure}[b]{\columnwidth}
        \centering
        \includegraphics[width=0.4\columnwidth]{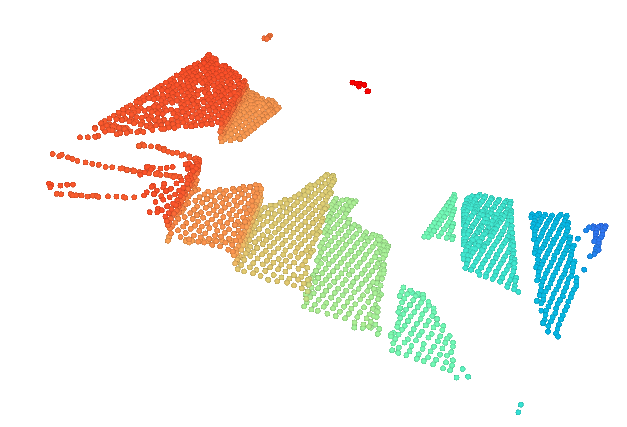}
        \hspace{0.1\columnwidth}
        \includegraphics[width=0.4\columnwidth]{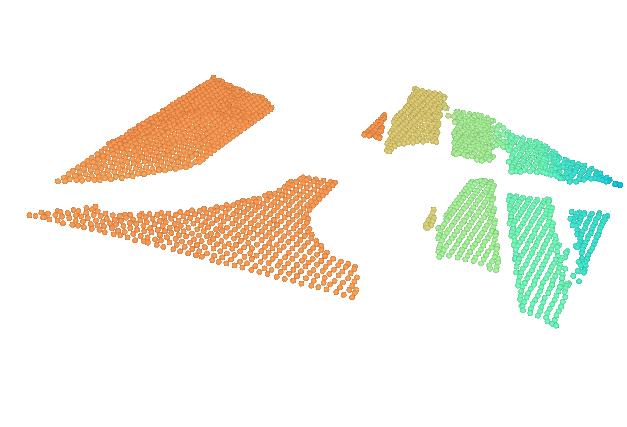}
        \caption{Measurement.}
        \label{fig:hoengg_meas}
    \end{subfigure}
    \begin{subfigure}[b]{\columnwidth}
        \centering
        \includegraphics[width=0.4\columnwidth]{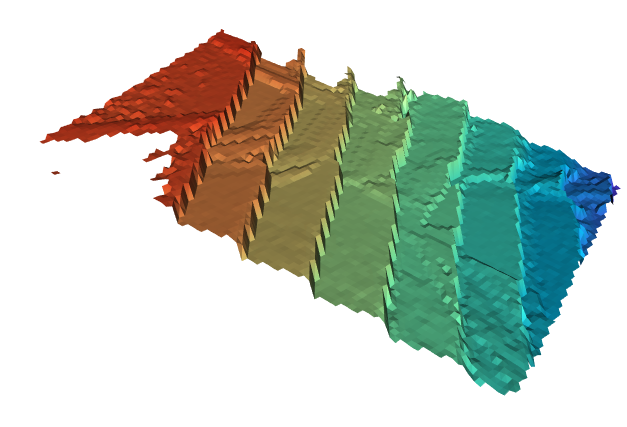}
        \hspace{0.1\columnwidth}
        \includegraphics[width=0.4\columnwidth]{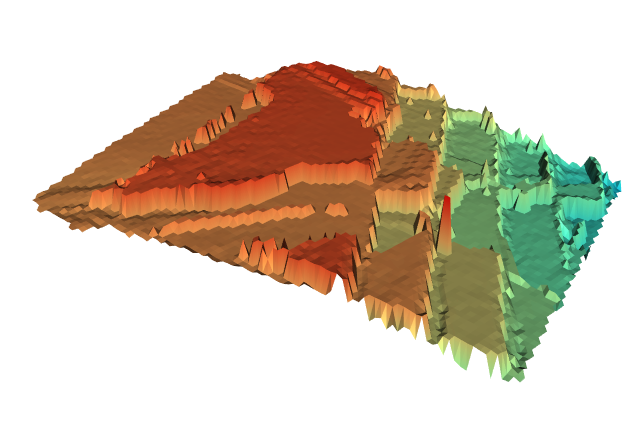}
        \caption{Elevation Mapping \cite{Fankhauser2018ProbabilisticTerrainMapping}.}
        \label{fig:hoengg_em}
     \end{subfigure}
    \begin{subfigure}[b]{\columnwidth}
        \centering
        \includegraphics[width=0.4\columnwidth]{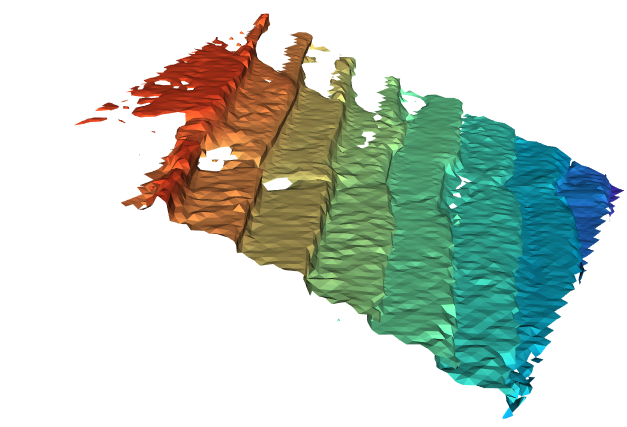}
        \hspace{0.1\columnwidth}
        \includegraphics[width=0.4\columnwidth]{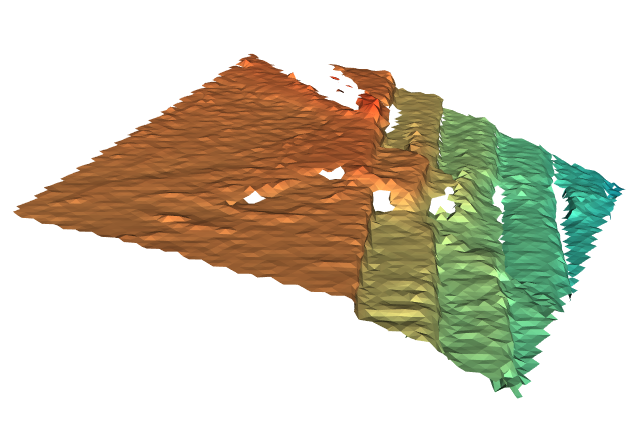}
        \caption{Voxblox \cite{oleynikova2017voxblox}.}
        \label{fig:hoengg_voxblox}
     \end{subfigure}
    \begin{subfigure}[b]{\columnwidth}
        \centering
        \includegraphics[width=0.4\columnwidth]{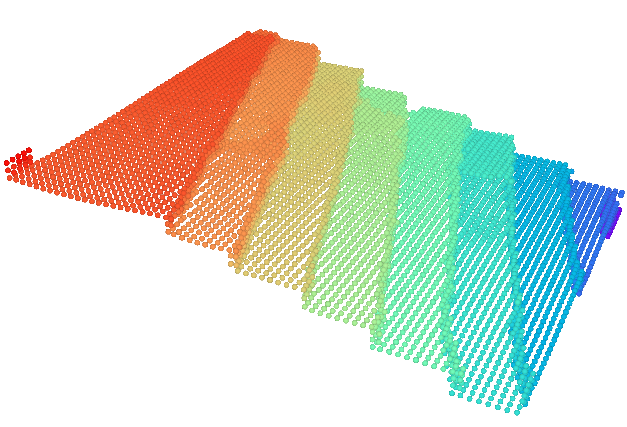}
        \hspace{0.1\columnwidth}
        \includegraphics[width=0.4\columnwidth]{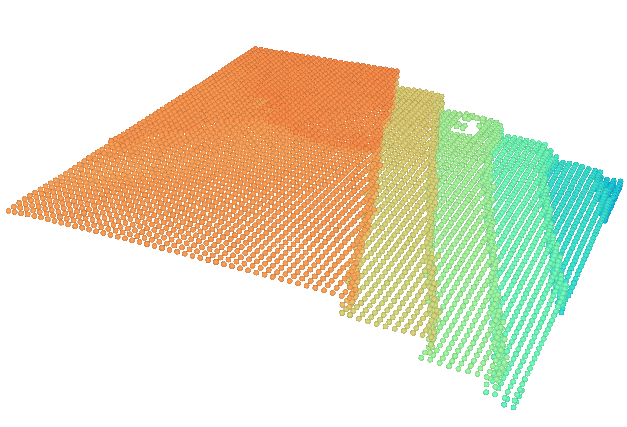}
        \caption{Our approach.}
        \label{fig:hoengg_ours}
     \end{subfigure}
    \begin{subfigure}[b]{\columnwidth}
        \centering
        \includegraphics[width=0.4\columnwidth]{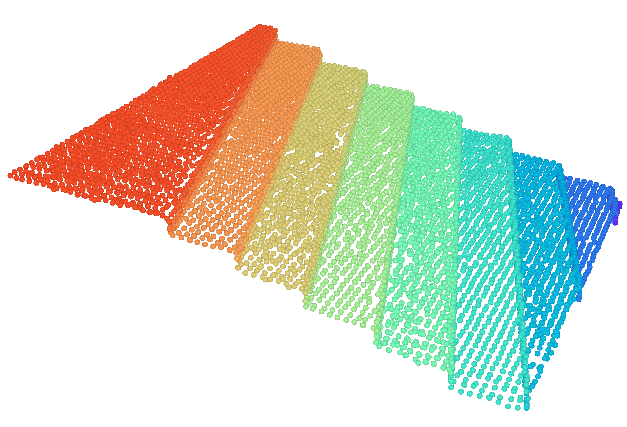}
        \hspace{0.1\columnwidth}
        \includegraphics[width=0.4\columnwidth]{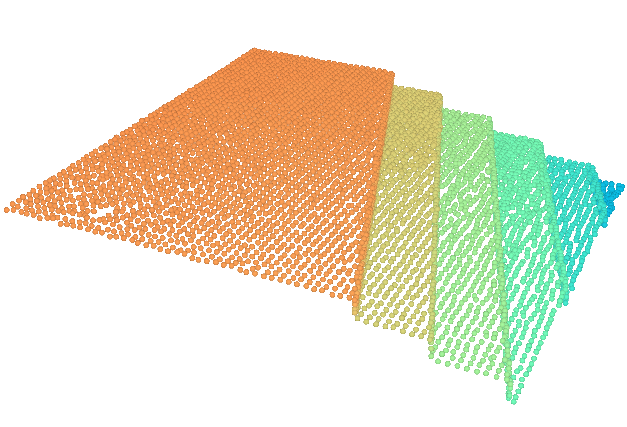}
        \caption{Ground truth.}
        \label{fig:hoengg_gt}
    \end{subfigure}
    \caption{Comparison of the maps along a trajectory with heavy state estimator drift. During the drift, our approach can detect the mismatch between the previous estimate and the current measurements and immediately aligns the map with the shifted measurements. While Elevation Mapping \cite{Fankhauser2018ProbabilisticTerrainMapping} cannot handle the drift, Voxblox \cite{oleynikova2017voxblox} can correct it reasonably well, except immediately below the robot.}
    \label{fig:hoengg}
    \vspace{-4mm}
\end{figure}

\begin{table}[t]
    \centering
    \vspace{2mm}
    \caption{Comparison of the different approaches on the box and stairs data sets.}
    \begin{tabular}{c | c c c c c}
        \hline
        { } & {Method} & \makecell{Pr. [$\%$]} & \makecell{Re. [$\%$]} & \makecell{F1 [$\%$] } & \makecell{MAE [$\SI{}{\centi\meter}$] } \\
        \hline
        \multirow{4}*{\rotatebox[origin=c]{90}{\textbf{Stairs}}} & {Measurement} & $87.7$ & $50.7$ & $64.0$ & $0.64$ \\
                                       & E.M. \cite{Fankhauser2018ProbabilisticTerrainMapping} &  $73.2$ & $79.8$ & $76.3$ & $2.1$ \\
                                       & Voxblox \cite{oleynikova2017voxblox} & $76.3$ & $72.3$ & $73.5$ & $1.6$ \\
                                       & Ours & $\mathbf{86.0}$ & $\mathbf{89.9}$ & $\mathbf{88.9}$ & $\mathbf{0.8}$ \\


        \hline
        \multirow{4}*{\rotatebox[origin=c]{90}{\textbf{Box}}} & {Measurement} & $80.8$ & $61.6$ & $69.8$ & $1.2$ \\
                                        & E.M. \cite{Fankhauser2018ProbabilisticTerrainMapping} & $72.2$ & $80.0$ & $75.9$ & $1.7$ \\
                                        & Voxblox \cite{oleynikova2017voxblox} & $74.0$ & $77.9$ & $75.8$ & $1.4$ \\
                                        & Ours & $\mathbf{84.8}$ & $\mathbf{84.9}$ & $\mathbf{84.8}$ & $\mathbf{1.0}$ \\
        \hline
    \end{tabular}
    \label{tab:comparison}
    \vspace{-4mm}
\end{table}

We evaluate the approach on the real robot in different scenarios and compare the results against the two classical baselines. \modif{Both of these methods are computationally efficient and are able to process the data at the same rate as the point cloud measurements are incoming, i.e. \SI{15}{\hertz}.} Since these approaches produce meshes, we sample dense point clouds from their terrain estimates and use these for comparison. 
The ground truth point clouds are generated using the accurate BLK2GO LiDAR scanner. We perform ICP registration between each measurement frame and the BLK2GO map, which accurately estimates the robot's pose relative to the ground truth.
In the \emph{Stairs} experiment, the robot walks on stairs of various sizes and textures under different illumination conditions. 
In the \emph{Box} experiment, the robot walks in a scene with a large box in bright conditions. The wooden surface of the box is challenging for the depth cameras and parts of it are sometimes invisible. 


The qualitative results of one of the trajectories in the \emph{Stairs} experiment are shown in Fig.~\ref{fig:hoengg}. The depth cameras provide good measurements, and the resulting maps produce meaningful results for all three approaches. The reconstructed stairs seem slightly curved on the map borders for all three approaches. This is because of the noise on the edges of the steps, which is stronger on the map's borders, see Fig.~\ref{fig:hoengg_meas}. The approaches have difficulties correctly identifying the end of the step in these regions. Fig.~\ref{fig:hoengg_ours} shows that our approach can contain small holes in the reconstruction. This is due to the pruning formulation and because we do not explicitly train the pipeline to produce a water-tight output. However, this does not impact the performance of our controller since the holes are very local. As the robot walks up, the state estimator drifts down by \SI{7}{\centi\meter} just after $t=\SI{2.76}{\second}$. This can be seen by comparing the two maps in Fig.~\ref{fig:hoengg_gt}. The top surface sinks and changes from red to orange, while it should stay the same color since the maps are expressed in the world frame. 
Our approach detects the discrepancy in height between the previous map and the current measurement and immediately shifts down the whole map estimate (Fig.~\ref{fig:hoengg_ours}). Elevation mapping is not able to cope with the drift and updates the map at the previous height with the new measurements (Fig.~\ref{fig:hoengg_em}), resulting in an uneven map. On the other hand, while Voxblox still produces an erroneous reconstruction just below the robot, it is capable of handling the drift and produces a better surface around the robot (Fig.~\ref{fig:hoengg_voxblox}). 

The quantitative evaluation of the trajectories is reported in Tab.~\ref{tab:comparison}.
Our method outperforms the other baselines by around 10\% in most metrics. The method not only detects more points correctly, but the mean absolute error of the reconstruction is also lower. 
While the mean absolute error does not seem high for Elevation Mapping, the regions with offsets in the map, such as in Fig.\ref{fig:hoengg_em} are detrimental for the locomotion policy, as we show in the next subsection. 
The \emph{Box} data-set is more challenging. The box has a reflective wooden surface and is sometimes invisible to the cameras. Moreover, the jerkier motions of the base represent a challenge for all three approaches. Due to the missing points on the box, it slightly changes dimension over time, see the supplementary video. 

\modif{Note that our local method compensates for the drift by implicitly aligning the measurements with the map. The map's frame will drift with the robot and not be consistent with the ground truth global frame. However, this does not matter for locomotion, since the map will be correct in the robot's reference frame.}




\subsection{\modif{Locomotion in Structured Terrains}}
\modif{The previous experiments show that our module can be used by the control policy to overcome urban environments. Compared to Elevation Mapping, the map is cleaner resulting in smoother motions. This can be seen in the video in the supplementary materials, where the robot using the baseline map moves erratically on stairs and the joints shake more. Since the policy is trained with noise in the height measurements, the robot is capable of recovering from the instabilities, but it comes at the cost of worse tracking performance. This makes it challenging for autonomous deployment.}

\begin{figure}[t]
    \vspace{2mm}
    \centering
    \includegraphics[width=0.99\columnwidth]{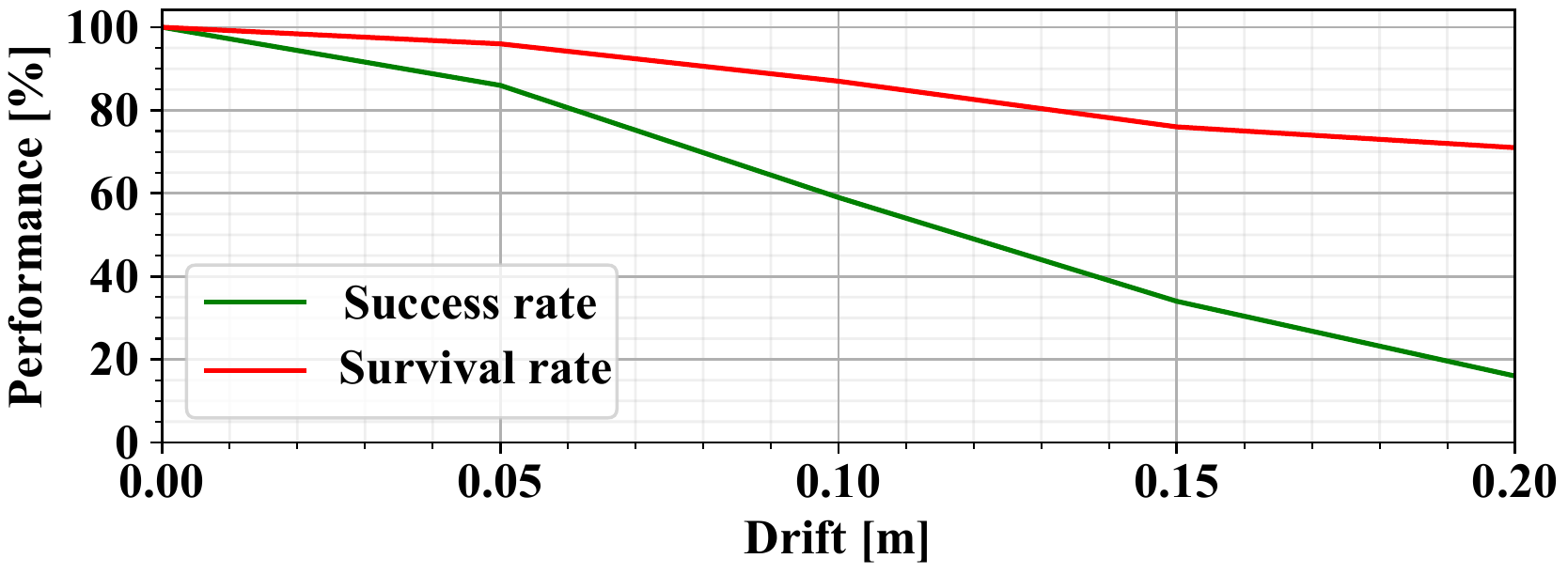}
    \caption{\modif{Performance on a target reaching task in simulation. We compute the average over 4000 trajectories. While the majority of robots do not fall for higher drifts, the tracking performance decreases and the robots fail to reach the target within the allocated time.}}
    \label{fig:result_sim}
    \vspace{-4mm}
\end{figure}

\modif{We assess the impact of the map quality on the policy in simulation on a target reaching task for varying amounts of drift. The robot has to cross a challenging terrain within \SI{7}{\second}, and to successfully complete the task, it has to move fast and place the footholds correctly. 
To model the drift that is varied from \SI{0}{\centi\meter} to \SI{20}{\centi\meter}, bumps of that height are randomly placed in the measured terrain. This results in a map that is similar to what we witness on the robot with the other mapping approaches. Fig.~\ref{fig:result_sim} shows that while the survival rate only decreases to 70\%, the quality of the map highly affects the tracking performance and thus the success rate of the task. For a drift of \SI{20}{\centi\meter}, fewer than 20\% of the robots manage to reach the target on time.
Additionally, compared to the drift-free case, the robot collides three times more often the knees with the ground with drifts of more than \SI{10}{\centi\meter}.} 

\modif{We deploy our map on a perceptive rough terrain model-based controller \cite{grandia2022perceptive} to show the generality of our module. This method computes feasible footholds on the terrain and tracks them using a model predictive controller. There, having an accurate map is crucial, because an erroneous foothold results in an unexpected trajectory that is hard to recover from, especially on stairs. Drifts in the map often result in failure, even when turning around on the spot in flat terrain, as can be seen in the supplementary video. We show in the video that using our map, the controller is capable of overcoming structured terrains.}

\section{CONCLUSION}

In this work, we presented a terrain representation module that uses noisy and occluded point cloud measurements on a robot to reconstruct the scene faithfully in the robot's vicinity. Our experiments in simulation and on the real robot show that it can successfully accumulate evidence spatially and temporally to build an accurate estimate of the terrain in structured environments. The robot's learning-based and model-based locomotion controllers are able to use our map to walk in an urban setting. 

The current approach still has some limitations that have to be tackled. 
The network can cope with dynamic obstacles reasonably well due to the augmentation with random distractors during training. However, some of the points of the dynamic elements remain in the reconstruction and only vanish slowly. 
A solution would be to create a data-set containing dynamic obstacles, or use a better noise model.

Also, while the reconstruction is clean most of the time, the depth cameras produce many outliers on the edges of objects. Since that noise is difficult to reproduce in simulation, the data-set could be extended with data from the real robot to make the pipeline ready for full autonomous deployment. Moreover, as mentioned in the introduction, the unstructured setting is even more challenging. Using data from the robot in such missions is probably key.

\modif{In the future, we would like to show the full potential of our method and make the robot walk below tables or climb higher obstacles, which would not be possible with other approaches such as Elevation Mapping. The strong prior induced by the training data could be used to reconstruct the top surface of higher obstacles which might not be visible to the sensor. Also, combining the point cloud data with proprioceptive data such as the position of the feet on contact might be useful when walking on compliant ground such as snow to find out the real ground height.
Finally, we would like to explore the direct use of the latent representation of our module by the policy. This could speed up the pipeline by avoiding the decoding stage and give bigger picture of the scene to the policy.}

\bibliographystyle{IEEEtran}
\bibliography{IEEEabrv, bibliography}

\begin{thebibliography}{10}
\providecommand{\url}[1]{#1}
\csname url@rmstyle\endcsname
\providecommand{\newblock}{\relax}
\providecommand{\bibinfo}[2]{#2}
\providecommand\BIBentrySTDinterwordspacing{\spaceskip=0pt\relax}
\providecommand\BIBentryALTinterwordstretchfactor{4}
\providecommand\BIBentryALTinterwordspacing{\spaceskip=\fontdimen2\font plus
\BIBentryALTinterwordstretchfactor\fontdimen3\font minus
  \fontdimen4\font\relax}
\providecommand\BIBforeignlanguage[2]{{%
\expandafter\ifx\csname l@#1\endcsname\relax
\typeout{** WARNING: IEEEtran.bst: No hyphenation pattern has been}%
\typeout{** loaded for the language `#1'. Using the pattern for}%
\typeout{** the default language instead.}%
\else
\language=\csname l@#1\endcsname
\fi
#2}}

\bibitem{belter2012estimating}
D.~Belter, P.~{\L}abcki, and P.~Skrzypczy{\'n}ski, ``Estimating terrain
  elevation maps from sparse and uncertain multi-sensor data,'' in \emph{2012
  IEEE International Conference on Robotics and Biomimetics (ROBIO)}.\hskip 1em
  plus 0.5em minus 0.4em\relax IEEE, 2012, pp. 715--722.

\bibitem{roennau2010six}
A.~Roennau, T.~Kerscher, M.~Ziegenmeyer, J.~Zoellner, and R.~Dillmann,
  ``Six-legged walking in rough terrain based on foot point planning,'' in
  \emph{Mobile robotics: solutions and challenges}.\hskip 1em plus 0.5em minus
  0.4em\relax World Scientific, 2010, pp. 591--598.

\bibitem{Fankhauser2018ProbabilisticTerrainMapping}
P.~Fankhauser, M.~Bloesch, and M.~Hutter, ``Probabilistic terrain mapping for
  mobile robots with uncertain localization,'' \emph{IEEE Robotics and
  Automation Letters (RA-L)}, vol.~3, no.~4, pp. 3019--3026, 2018.

\bibitem{mastalli2017trajectory}
C.~Mastalli \emph{et~al.}, ``Trajectory and foothold optimization using
  low-dimensional models for rough terrain locomotion,'' in \emph{2017 IEEE
  International Conference on Robotics and Automation (ICRA)}.\hskip 1em plus
  0.5em minus 0.4em\relax IEEE, 2017, pp. 1096--1103.

\bibitem{schulman2017ppo}
J.~Schulman, F.~Wolski, P.~Dhariwal, A.~Radford, and O.~Klimov, ``Proximal
  policy optimization algorithms,'' \emph{CoRR}, vol. abs/1707.06347, 2017.

\bibitem{miki2021locomotion}
T.~Miki, J.~Lee, J.~Hwangbo, L.~Wellhausen, V.~Koltun, and M.~Hutter,
  ``Learning robust perceptive locomotion for quadrupedal robots in the wild,''
  \emph{Science Robotics}, vol.~7, no.~62, p. eabk2822, 2022.

\bibitem{oleynikova2017voxblox}
H.~Oleynikova, Z.~Taylor, M.~Fehr, R.~Siegwart, and J.~Nieto, ``Voxblox:
  Incremental 3d euclidean signed distance fields for on-board mav planning,''
  in \emph{IEEE/RSJ International Conference on Intelligent Robots and Systems
  (IROS)}, 2017.

\bibitem{IsaacGym}
V.~Makoviychuk \emph{et~al.}, ``Isaac gym: High performance {GPU} based physics
  simulation for robot learning,'' in \emph{Conference on Neural Information
  Processing Systems (NeurIPS) Datasets and Benchmarks Track}, 2021.

\bibitem{kim2019loco}
D.~Kim, J.~D. Carlo, B.~Katz, G.~Bledt, and S.~Kim, ``Highly dynamic quadruped
  locomotion via whole-body impulse control and model predictive control,''
  \emph{CoRR}, vol. abs/1909.06586, 2019.

\bibitem{tan2018simtoreal}
J.~Tan, T.~Zhang, E.~Coumans, A.~Iscen, Y.~Bai, D.~Hafner, S.~Bohez, and
  V.~Vanhoucke, ``Sim-to-real: Learning agile locomotion for quadruped
  robots,'' in \emph{Proceedings of Robotics: Science and Systems}, Pittsburgh,
  Pennsylvania, June 2018.

\bibitem{hwangbo2019RL}
J.~Hwangbo, J.~Lee, A.~Dosovitskiy, D.~Bellicoso, V.~Tsounis, V.~Koltun, and
  M.~Hutter, ``Learning agile and dynamic motor skills for legged robots,''
  \emph{Science Robotics}, vol.~4, no.~26, 2019.

\bibitem{lee2020locomotion}
J.~Lee, J.~Hwangbo, L.~Wellhausen, V.~Koltun, and M.~Hutter, ``Learning
  quadrupedal locomotion over challenging terrain,'' \emph{Science Robotics},
  vol.~5, no.~47, p. eabc5986, 2020.

\bibitem{siekmann2021Cassie}
J.~Siekmann, K.~Green, J.~Warila, A.~Fern, and J.~Hurst, ``{Blind Bipedal Stair
  Traversal via Sim-to-Real Reinforcement Learning},'' in \emph{Proceedings of
  Robotics: Science and Systems}, July 2021.

\bibitem{rudin2021learning}
N.~Rudin, D.~Hoeller, P.~Reist, and M.~Hutter, ``Learning to walk in minutes
  using massively parallel deep reinforcement learning,'' in \emph{5th Annual
  Conference on Robot Learning}, 2021.

\bibitem{gangapurwala2020RLOCTL}
S.~Gangapurwala, M.~Geisert, R.~Orsolino, M.~F. Fallon, and I.~Havoutis,
  ``Rloc: Terrain-aware legged locomotion using reinforcement learning and
  optimal control,'' \emph{ArXiv}, vol. abs/2012.03094, 2020.

\bibitem{yang2022learning}
R.~Yang, M.~Zhang, N.~Hansen, H.~Xu, and X.~Wang, ``Learning vision-guided
  quadrupedal locomotion end-to-end with cross-modal transformers,'' in
  \emph{International Conference on Learning Representations}, 2022.

\bibitem{klein2007parallel}
G.~Klein and D.~Murray, ``Parallel tracking and mapping for small ar
  workspaces,'' in \emph{2007 6th IEEE and ACM international symposium on mixed
  and augmented reality}.\hskip 1em plus 0.5em minus 0.4em\relax IEEE, 2007,
  pp. 225--234.

\bibitem{forster2014svo}
C.~Forster, M.~Pizzoli, and D.~Scaramuzza, ``Svo: Fast semi-direct monocular
  visual odometry,'' in \emph{2014 IEEE international conference on robotics
  and automation (ICRA)}.\hskip 1em plus 0.5em minus 0.4em\relax IEEE, 2014,
  pp. 15--22.

\bibitem{engel2014lsd}
J.~Engel, T.~Sch{\"o}ps, and D.~Cremers, ``Lsd-slam: Large-scale direct
  monocular slam,'' in \emph{European conference on computer vision}.\hskip 1em
  plus 0.5em minus 0.4em\relax Springer, 2014, pp. 834--849.

\bibitem{hornung13octomap}
A.~Hornung, K.~M. Wurm, M.~Bennewitz, C.~Stachniss, and W.~Burgard,
  ``{OctoMap}: An efficient probabilistic {3D} mapping framework based on
  octrees,'' \emph{Autonomous Robots}, 2013.

\bibitem{kim2020loco}
D.~Kim, D.~Carballo, J.~Di~Carlo, B.~Katz, G.~Bledt, B.~Lim, and S.~Kim,
  ``Vision aided dynamic exploration of unstructured terrain with a small-scale
  quadruped robot,'' in \emph{2020 IEEE International Conference on Robotics
  and Automation (ICRA)}, 2020, pp. 2464--2470.

\bibitem{chavez2018traversability}
R.~O. Chavez-Garcia, J.~Guzzi, L.~M. Gambardella, and A.~Giusti, ``Learning
  ground traversability from simulations,'' \emph{IEEE Robotics and Automation
  Letters}, vol.~3, no.~3, pp. 1695--1702, 2018.

\bibitem{9676411}
M.~Stölzle, T.~Miki, L.~Gerdes, M.~Azkarate, and M.~Hutter, ``Reconstructing
  occluded elevation information in terrain maps with self-supervised
  learning,'' \emph{IEEE Robotics and Automation Letters}, vol.~7, no.~2, pp.
  1697--1704, 2022.

\bibitem{niessner2013real}
M.~Nie{\ss}ner, M.~Zollh{\"o}fer, S.~Izadi, and M.~Stamminger, ``Real-time 3d
  reconstruction at scale using voxel hashing,'' \emph{ACM Transactions on
  Graphics (ToG)}, vol.~32, no.~6, pp. 1--11, 2013.

\bibitem{song2017semantic}
S.~Song, F.~Yu, A.~Zeng, A.~X. Chang, M.~Savva, and T.~Funkhouser, ``Semantic
  scene completion from a single depth image,'' in \emph{Proceedings of the
  IEEE Conference on Computer Vision and Pattern Recognition}, 2017, pp.
  1746--1754.

\bibitem{dai2018scancomplete}
A.~Dai, D.~Ritchie, M.~Bokeloh, S.~Reed, J.~Sturm, and M.~Nie{\ss}ner,
  ``Scancomplete: Large-scale scene completion and semantic segmentation for 3d
  scans,'' in \emph{Proceedings of the IEEE Conference on Computer Vision and
  Pattern Recognition}, 2018, pp. 4578--4587.

\bibitem{Choe2021VolumeFusion}
J.~Choe, S.~Im, F.~Rameau, M.~Kang, and I.~S. Kweon, ``Volumefusion: Deep depth
  fusion for 3d scene reconstruction,'' in \emph{Proceedings of the IEEE/CVF
  International Conference on Computer Vision (ICCV)}, 2021, pp.
  16\,086--16\,095.

\bibitem{sun2021neuralrecon}
J.~Sun, Y.~Xie, L.~Chen, X.~Zhou, and H.~Bao, ``Neuralrecon: Real-time coherent
  3d reconstruction from monocular video,'' in \emph{Proceedings of the
  IEEE/CVF Conference on Computer Vision and Pattern Recognition}, 2021, pp.
  15\,598--15\,607.

\bibitem{gwak2020gsdn}
J.~Gwak, C.~B. Choy, and S.~Savarese, ``Generative sparse detection networks
  for 3d single-shot object detection,'' in \emph{European conference on
  computer vision}, 2020.

\bibitem{ronneberger2015Unet}
O.~Ronneberger, P.~Fischer, and T.~Brox, ``U-net: Convolutional networks for
  biomedical image segmentation,'' in \emph{Medical Image Computing and
  Computer-Assisted Intervention -- MICCAI 2015}, 2015, pp. 234--241.

\bibitem{choy20194d}
C.~Choy, J.~Gwak, and S.~Savarese, ``4d spatio-temporal convnets: Minkowski
  convolutional neural networks,'' in \emph{Proceedings of the IEEE Conference
  on Computer Vision and Pattern Recognition}, 2019, pp. 3075--3084.

\bibitem{grandia2022perceptive}
R.~Grandia, F.~Jenelten, S.~Yang, F.~Farshidian, and M.~Hutter, ``Perceptive
  locomotion through nonlinear model predictive control,'' \emph{(submitted to)
  IEEE Transactions on Robotics}, 2022.

\end{thebibliography}



\end{document}